\documentclass[runningheads]{llncs}
\usepackage{graphicx}
\usepackage{booktabs}
\usepackage[doi=false,isbn=false,url=false,eprint=false,sorting=none]{biblatex}
\usepackage{amsmath}
\usepackage{svg}
\usepackage{cancel}
\usepackage{tabularx}
\usepackage{amsfonts}

\newcommand{\beginsupplement}{%
        \setcounter{table}{0}
        \renewcommand{\thetable}{S\arabic{table}}%
        \setcounter{figure}{0}
        \renewcommand{\thefigure}{S\arabic{figure}}%
     }
     
\makeatletter
\def\thanks#1{\protected@xdef\@thanks{\@thanks
        \protect\footnotetext{#1}}}
\makeatother

\begin{document}

\title{Deep Orthogonal Fusion: Multimodal Prognostic Biomarker Discovery Integrating Radiology, Pathology, Genomic, and Clinical Data\thanks{Accepted for presentation at MICCAI 2021.}}

\titlerunning{Deep Orthogonal Fusion for Multimodal Prognostic Biomarker Discovery}
\author{Nathaniel Braman, Jacob W. H. Gordon, Emery T. Goossens, Caleb Willis, Martin C. Stumpe, Jagadish Venkataraman}

\authorrunning{N. Braman et al.}
\institute{Tempus Labs, Inc., Chicago, IL, USA\\ 
\email{\{nathaniel.braman, jagadish.venkataraman\}@tempus.com}\\
\url{https://www.tempus.com/}}

\maketitle              %

\begin{abstract}
Clinical decision-making in oncology involves multimodal data such as radiology scans, molecular profiling, histopathology slides, and clinical factors. Despite the importance of these modalities individually, no deep learning framework to date has combined them all to predict patient prognosis. Here, we predict the overall survival (OS) of glioma patients from diverse multimodal data with a Deep Orthogonal Fusion (DOF) model. The model learns to combine information from multiparametric MRI exams, biopsy-based modalities (such as H\&E slide images and/or DNA sequencing), and clinical variables into a comprehensive multimodal risk score. The model learns to combine embeddings from each modality via attention-gated tensor fusion. To maximize the information gleaned from each modality, we introduce a multimodal orthogonalization (MMO) loss term that increases model performance by incentivizing constituent embeddings to be more complementary. DOF predicts OS in glioma patients with a median C-index of 0.788 ± 0.067, significantly outperforming (p=0.023) the best performing unimodal model with a median C-index of 0.718 ± 0.064. The prognostic model significantly stratifies glioma patients by OS within clinical subsets, adding further granularity to prognostic clinical grading and molecular subtyping.
\end{abstract}

\section{Introduction}

Cancer diagnosis and treatment plans are guided by multiple streams of data acquired from several modalities, such as radiology scans, molecular profiling, histology slides, and clinical variables. Each characterizes unique aspects of tumor biology and, collectively, they help clinicians understand patient prognosis and assess therapeutic options. Advances in molecular profiling techniques have enabled the discovery of prognostic gene signatures, bringing precision medicine to the forefront of clinical practice \cite{el-deiry_current_2019}. More recently, computational techniques in the field of radiology have identified potential imaging-based phenotypes of treatment response and patient survival. Such approaches leverage large sets of explicitly designed image features (commonly known as radiomics \cite{gillies_radiomics:_2015}) or entail the novel discovery of image patterns by optimizing highly parameterized deep learning models such as convolutional neural networks (CNN) \cite{saba_present_2019} for prediction. Along similar lines, the digitization of histopathology slides has opened new avenues for tissue-based assays that can stratify patients by risk from H\&E slide images alone \cite{skrede_deep_2020}. Given the complementary nature of these various modalities in comprehensive clinical assessment, we hypothesize that their combination in a rigorous machine learning framework may predict patient outcomes more robustly than qualitative clinical assessment or unimodal strategies. 

Glioma is an intuitive candidate for deep learning-based multimodal biomarkers owing to the presence of well-characterized prognostic information across modalities \cite{louis_2016_2016}, as well as its severity \cite{siegel_cancer_2017}.  Gliomas can be subdivided by their malignancy into histological grades II-IV \cite{louis_2016_2016}. Grades differ in their morphologic and molecular heterogeneity \cite{olar_using_2014}, which correspond to treatment resistance and short-term recurrence \cite{stupp_radiotherapy_2005,parker_intratumoral_2016}. Quantitative analysis of glioma \cite{bae_radiomic_2018} and its tumor habitat \cite{prasanna_radiomic_2017} on MRI has demonstrated strong prognostic potential, as well as complex interactions with genotype \cite{beig_radiogenomic-based_2020} and clinical variables \cite{beig_sexually_2021}.

Most deep multimodal prediction strategies to date have focused on the fusion of biopsy-based modalities \cite{chen_pathomic_2019,mobadersany_predicting_2018,cheerla_deep_2019}. For instance, previous work integrating molecular data with pathology analysis via CNN or graph convolutional neural networks (GCN) has shown that a deep, multimodal approach improves prognosis prediction in glioma patients  \cite{chen_pathomic_2019,mobadersany_predicting_2018}. Likewise, Cheerla et al. integrated histology, clinical, and sequencing data across cancer types by condensing each to a correlated prognostic feature representation \cite{cheerla_deep_2019}. Multimodal research involving radiology has been predominantly correlative in nature \cite{beig_sexually_2021,beig_radiogenomic-based_2020}. Some have explored late-stage fusion approaches combining feature-based representations from radiology with similar pathology \cite{vaidya_raptomics_2018} or genomic features \cite{subramanian_multimodal_2020} to predict recurrence. While promising, these strategies rely on hand-crafted feature sets and simple multimodal classifiers that likely limit their ability to learn complex prognostic interactions between modalities and realize the full additive benefit of integrating diverse clinical modalities. 

To our knowledge, no study to date has combined radiology, pathology, and genomic data within a single deep learning framework for outcome prediction or patient stratification. Doing so requires overcoming several challenges. First, owing to the difficulty of assembling multimodal datasets with corresponding outcome data in large quantities, fusion schemes must be highly data efficient in learning complex multimodal interactions. Second, the presence of strongly correlated prognostic signals between modalities \cite{cheerla_deep_2019} can create redundancy and hinder model performance.  

\begin{figure}
\centering
\includegraphics[width=1\textwidth]{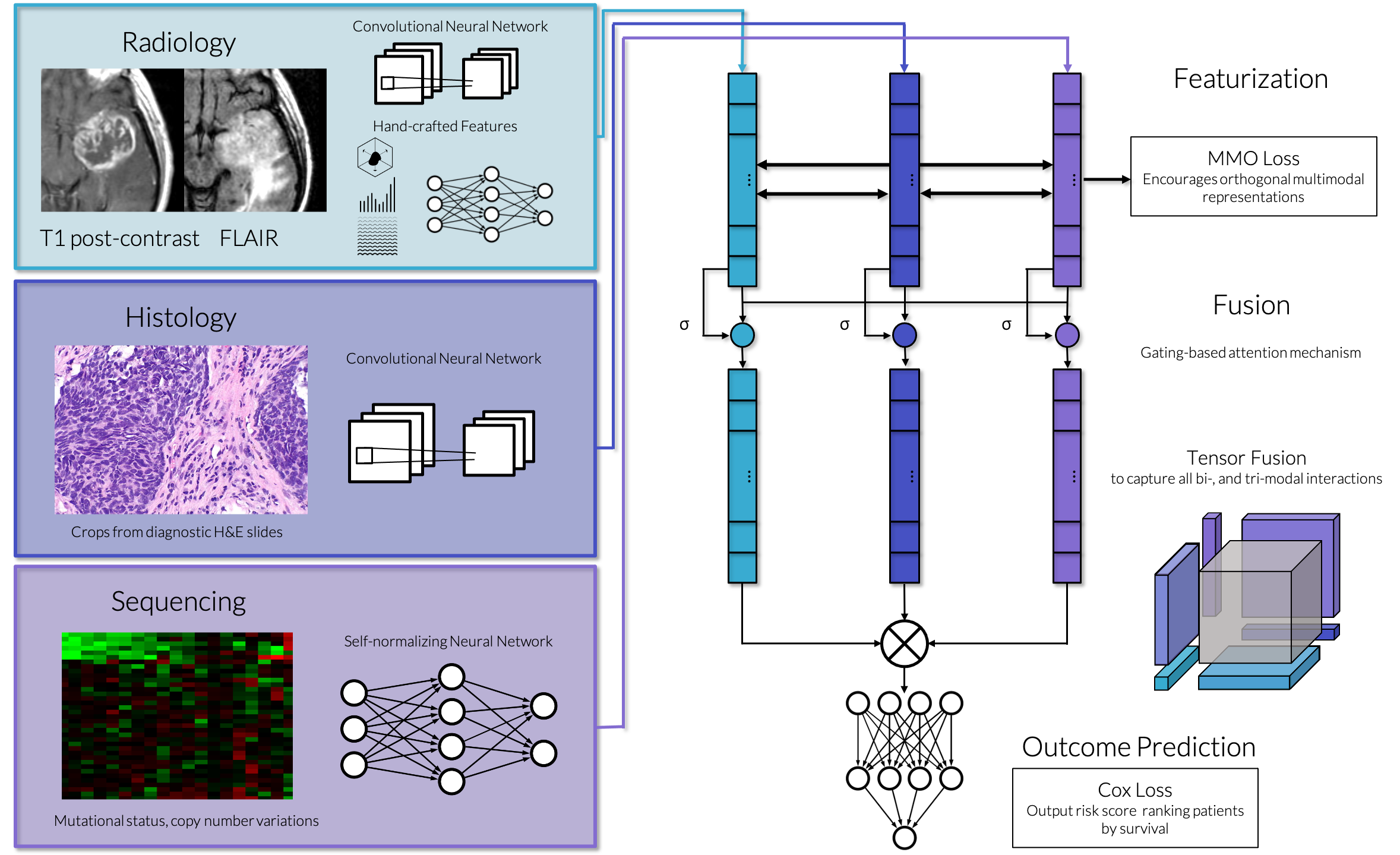}
\caption{DOF model architecture and training.} \label{fig1}
\end{figure}

In this paper, we introduce a deep learning framework that combines radiologic, histologic, genomic, and clinical data into a fused prognostic risk score. Using a novel technique referred to as Deep Orthogonal Fusion (DOF), we train models using a Multimodal Orthogonalization (MMO) loss function to maximize the independent contribution of each data modality, effectively improving predictive performance. Our approach, depicted in Fig. \ref{fig1}, first trains unimodal embeddings for overall survival (OS) prediction through a Cox partial likelihood loss function. Next, these embeddings are combined through an attention-gated tensor fusion to capture all possible interactions between each data modality. Fusion models are trained simultaneously to predict OS and minimize the correlation between unimodal embeddings. We emphasize the following contributions: 

\textbf{Deep Fusion of Radiology, Pathology, and Omics Data}: We present a powerful, data-efficient framework for combining oncologic data across modalities. Our approach enabled a previously unexplored deep integration of radiology with tissue-based modalities and clinical variables for patient risk stratification. This fusion model significantly improved upon unimodal deep learning models. In particular, we found that integrating radiology into deep multimodal models, which is under-explored in previous prognostic studies, conferred the single greatest performance increase. This finding suggests the presence of independent, complementary prognostic information between radiology and biopsy-based modalities and warrants their combination in future prognostic studies. 

\textbf{MMO}: To mitigate the effect of inherent correlations between data modalities, we present an MMO loss function that penalizes correlation between unimodal embeddings and encourages each to provide independent prognostic information. We find that this training scheme, which we call DOF, improves prediction by learning and fusing disentangled, prognostic representations from each modality. DOF was also found to outperform a fusion scheme that enforces correlated representations between modalities \cite{cheerla_deep_2019}, emphasizing that the dissimilarity of these clinical data streams is crucial to their collective strength. 

\textbf{Multi-parametric Radiology FeatureNet:} A neural network architecture that can fuse CNN-extracted deep features from local tumor regions on multiple image sequences (e.g., Gd-T1w and T2w-FLAIR scans) with global hand-crafted radiomics features extracted across the full 3D region-of-interest.

\textbf{Independent prognostic biomarker of OS in glioma patients:} Using 15-fold Monte Carlo cross-validation with a 20\% holdout test set, we evaluate deep fusion models to predict glioma prognosis. We compare this multimodal risk score with existing prognostic clinical subsets and biomarkers (grade, \textit{IDH} status) and investigate its prognostic value within these outcome-associated groups.

\section{Methodology}

Let $X$ be a training minibatch of data for $N$ patients, each containing $M$ modalities such that $X = [x_1, x_2, ... , x_M]$. For each modality $m$, $x_m$ includes data from for $N$ patients. $\Phi_m$ denotes a trainable unimodal network, which accepts $x_m$ and generates a deep embedding $h_m = \Phi_m(x_m) \in \mathbb{R}^{l_1 x N}$.

\subsection{Multimodal Fusion}
When $M>1$, we combine embeddings from each modality in a multimodal fusion network. For each $h_m$, an attention mechanism is applied to control its expressiveness  based on information from the other modalities. An additional fully connected layer results in $h_m^S$ of length $l_2$. Attention weights of length $l_2$ are obtained through a bilinear transformation of $h_m$ with all other embeddings (denoted as $H_{\cancel{m}}$), then applied to $h_m^S$ to yield the attention-gated embedding:
\begin{equation}
 h_m^* = a_m * h_m^S = \sigma (h_m^T * W_A * H_{\cancel{m}})* h_m^S.
\end{equation}

To capture all possible interactions between modalities, we combine attention-weighted embeddings through an outer product between modalities, known as tensor fusion \cite{zadeh_tensor_2017}. A value of 1 is also included in each vector, allowing for partial interactions between modalities and for the constituent unimodal embeddings to be retained. The output matrix
\begin{equation}
 F =  \begin{bmatrix} 1 \\
h_{1}^* \end{bmatrix} \otimes \begin{bmatrix} 1 \\ 
 h_{2}^* \end{bmatrix} \otimes ... \otimes \begin{bmatrix} 1 \\
h_{M}^* \end{bmatrix}
\end{equation} 
is an $M$-dimensional hypercube of all multimodal interactions with sides of length $l_2 + 1$. Fig. \ref{fig1} depicts $F$ for the fusion of radiology, pathology, and genomic data. It contains subregions corresponding to unaltered unimodal embeddings, pairwise fusions between 2 modalities, and trilinear fusion between all three of the modalities. A final set of fully connected layers, denoted by $\Phi_F$, is applied to tensor fusion features for a final fused embedding $h_F = \Phi_F (F)$.

\subsection{MMO Loss}
To address the shortcoming of multimodal models converging to correlated predictors, we introduce MMO loss. Inspired by Orthogonal Low-rank Embedding \cite{lezama_ole_2017}, we stipulate that unimodal embeddings preceding fusion should be orthogonal. This criterion enforces that each modality introduced contributes unique information to outcome prediction, rather than relying on signal redundancy between modalities. Each $\Phi_m$ is updated through MMO loss to yield embeddings that better complement other modalities. Let $ H \in \mathbb{R}^{l_1 x M*N}$ be the set of embeddings from all modalities. MMO loss is computed as
\begin{equation}
L_{MMO} = \frac{1}{M*N} \sum_{m=1}^M max(1, ||h_m||_*) - ||H||_*
\end{equation}
where $ || \cdot ||_*$ denotes the matrix nuclear norm (i.e., the sum of the matrix singular values). This loss is the difference between the sum of nuclear norms per embedding and the nuclear norm of all embeddings combined. It penalizes the scenario where the variance of two modalities separately is decreased when combined and minimized when all unimodal embeddings are fully orthogonal. The per-modality norm is bounded to a minimum of 1 to prevent the collapse of embedding features to zero.

\subsection{Cox Partial Likelihood Loss}
The final layer of each network, parameterized by $\beta$, is a fully connected layer with a single unit. This output functions as a Cox proportional hazards model using the deep embedding from the previous layer, $h$, as its covariates. This final layer's output, $\theta$, is the log hazard ratio, which is used as a risk score. The log hazard ratio for patient $i$ is denoted as $ \theta_i = h_i^T * \beta $.

We define the negative log likelihood $L_{pl}$ as our cost function 
\begin{equation}
L_{pl} = - \sum_{i:  E_i=1} \left ( \theta_i - log \sum_{j: t_i \geq  t_j} e^{\theta_j}    \right )
\end{equation}
where $t \in \mathbb{R}^{Nx1} $ indicates the time to date of last follow up. The event vector, $E\in\{0,1\}^{Nx1}$, equals 1 if  an event was observed (death) or 0 if a patient was censored (still alive) at time of last follow up. Each patient $i$ with an observed event is compared against all patients whose observation time was greater than or equal to $t_i$. Networks are trained using the final loss $L$ which is a linear combination of the two loss functions specified above
\begin{equation}
 L = L_{pl} + \gamma L_{MMO}
\end{equation}
where $\gamma$ is a scalar weighting the contribution of MMO loss relative to Cox partial likelihood loss. When training unimodal networks, $\gamma$ is always zero. Performance for various values of $\gamma$ are included in the Table \ref{tab:tabs4}.

\subsection{Modality-specific Networks for Outcome Prediction}
\textbf{Radiology: }A multiple-input CNN was designed to incorporate multiparametric MRI data and global lesion measurements, shown in Fig. \ref{figS1}. The backbone of the network is a VGG-19 CNN \cite{simonyan_very_2015} with batch normalization, substituting the final max pooling layer with a 4x4 adaptive average pooling. Two pre-trained \cite{deng_imagenet_2009} CNN branches  separately extract features from Gd-T1w and T2w-FLAIR images, which are then concatenated and passed through a fully connected layer. A third branch passes hand-crafted features (described in section 3) through a similar fully connected layer. Concatenated embeddings from all branches are fed to 2 additional fully connected layers. All fully connected layers preceding the final embedding layer have 128 units. 

\textbf{Histology, genomic, and clinical data:} We reused the models proposed in \cite{chen_pathomic_2019} - a pre-trained VGG-19 CNN with pretrained convolutional layers for Histology and a Self-Normalizing Neural Network (SNN) for genomic data. We also use this SNN for analysis of clinical data, which was not explored in \cite{chen_pathomic_2019}.

  \begin{figure}
 \centering
\includegraphics[width=.9\textwidth]{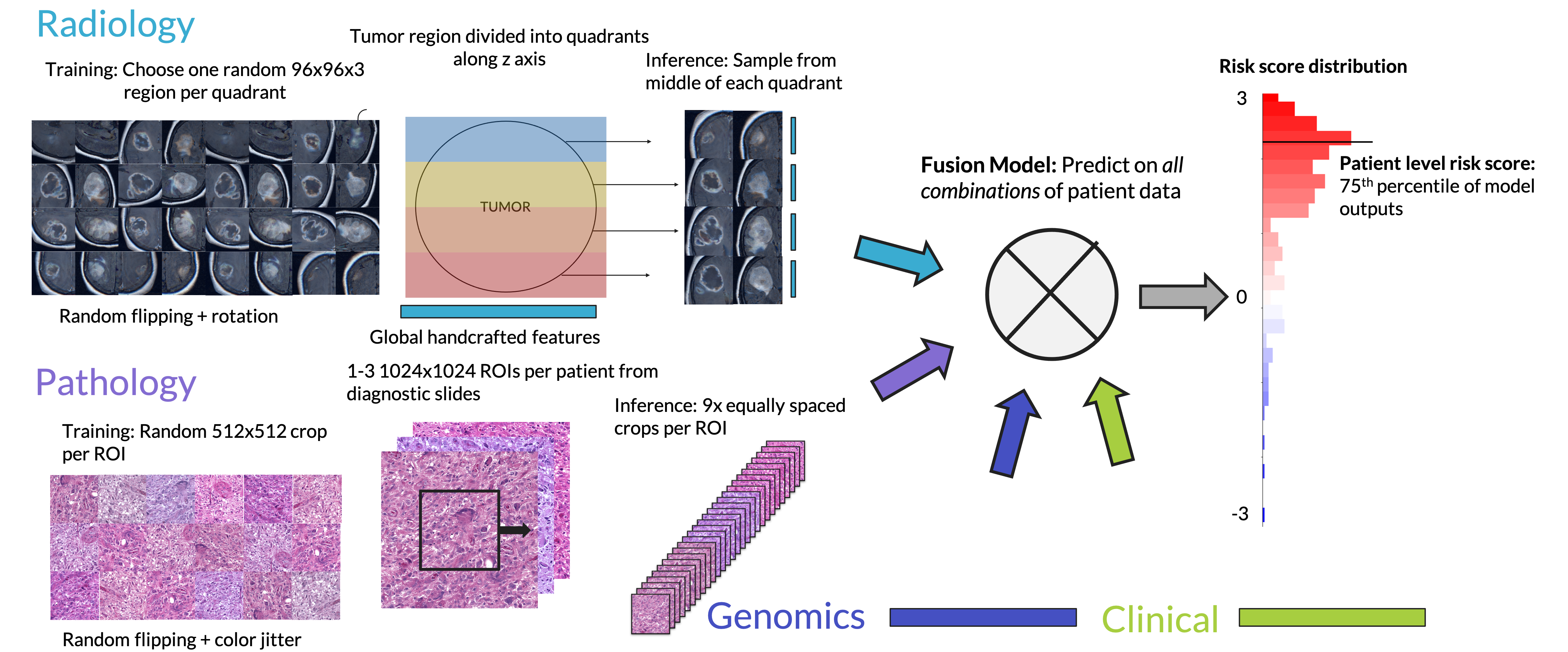}
\caption{Sampling multiple radiology \& pathology images for patient level risk scores.} \label{fig2}
\end{figure}

\section{Experimental Details}

\textbf{Radiology}: 176 patients (see patient selection in Fig. \ref{figS2}) with Gd-T1w and T2w-FLAIR scans from the TCGA-GBM \cite{scarpace_radiology_2016} and TCGA-LGG \cite{pedano_radiology_2016} studies were obtained from TCIA \cite{clark_cancer_2013} and annotated by 7 radiologists to delineate the enhancing lesion and edema region. Volumes were registered to the MNI-ICBM standardized brain atlas with 1 mm isotropic resolution, processed with N4 bias correction, and intensity normalized. 96x96x3 patches were generated from matching regions of Gd-T1w and T2w-FLAIR images within the enhancing lesion. For each patient, 4 samples were generated from four even quadrants of the tumor along the z-axis. Patch slice position was randomized in unimodal training and fixed to the middles of quadrants during inference and fusion network training. Nine features including size, shape, and intensity measures were extracted separately from Gd-T1w and T2w-FLAIR images, and summarized in three fashions for a total of 56 handcrafted features, listed in Table \ref{tab:tabs1}. 

\textbf{Pathology and Genomics:} We obtained 1024×1024 normalized regions-of-interest (ROIs) and DNA sequencing data curated by \cite{mobadersany_predicting_2018}. Each patient had 1-3 ROIs from diagnostic H\&E slides, totaling 372 images. DNA data consisted of 80 features including mutational status and copy number variation (Table \ref{tab:tabs2}).

\textbf{Clinical information: }14 clinical features were included into an SNN for the prediction of prognosis. The feature set included demographic and treatment details, as well as subjective histological subtype (see Table \ref{tab:tabs3}). 

\textbf{Implementation Details:} The embedding size for unimodal networks, $l_1$, was set to 32. Pre-fusion scaled embedding size, $l_2$, was 32 for $M$=2, 16 for $M$=3, and 8 for $M$=4. Post-fusion fully connected layers consisted of 128 units each. The final layer of each network had a single unit with sigmoid activation, but its outputs were rescaled between -3 and 3 to function as a prognostic risk score. Unimodal networks were trained for 50 epochs with linear learning rate decay, while multimodal networks were trained for 30 epochs with learning rate decay beginning at the 10th epoch. When training multimodal networks, the unimodal embedding layers were frozen for 5 epochs to train the fusion layers only, then unfrozen for joint training of embeddings and fusion layers. 

\textbf{Statistical Analysis: }All models were trained via 15-fold Monte Carlo cross-validation with 20\% holdout using the patient-level splits  provided in \cite{mobadersany_predicting_2018}. The primary performance metric was the median observed concordance index (C-index) across folds, a global metric of prognostic model discriminant power. We evaluated all possible combinations of a patient’s data (see sampling strategy in Fig. \ref{fig2}) and used the 75th percentile of predicted risk score as their overall prediction. C-indexes of the best-performing unimodal model and the DOF multimodal model were compared with a Mann-Whitney $U$ test \cite{mann_test_1947}. Binary low/high-risk groups were derived from the risk scores, where a risk score $>$0 corresponded to high risk. For Kaplan-Meier (KM) curves and calculation of hazard ratio (HR), patient-level risk scores were pooled across validation folds. 

\begin{table}[]
\centering
\caption{Median C-index of unimodal and fusion models with and without MMO loss. }
\label{tab:table1}
\begin{tabular}{@{}cccc@{}}
\toprule
\textbf{Group}  & \textbf{Model}    & \textbf{Cox Loss Only} & \textbf{With MMO Loss} \\ \midrule
Unimodal        & Rad               & 0.718 ± 0.064          & --                     \\
                & Path              & 0.715 ± 0.054          & --                     \\
                & Gen               & 0.716 ± 0.063          & --                     \\
                & Clin              & 0.702 ± 0.049          & --                     \\
Pairwise Fusion & Path+Gen          & 0.711 ± 0.055          & 0.752 ± 0.072          \\
                & Gen+Clin          & 0.702 ± 0.053          & 0.703 ± 0.052          \\
                & Rad+Gen           & 0.761 ± 0.071          & 0.766 ± 0.067          \\
                & Rad+Path          & 0.742 ± 0.067          & 0.752 ± 0.072          \\
                & Rad+Clin          & 0.746 ± 0.068          & 0.736 ± 0.067          \\
                & Path+Clin         & 0.696 ± 0.051          & 0.690 ± 0.043          \\
Triple Fusion   & Path+Gen+Clin     & 0.704 ± 0.059          & 0.720 ± 0.056          \\
                & Rad+Path+Clin     & 0.748 ± 0.067          & 0.741 ± 0.067          \\
                & Rad+Gen+Clin      & 0.754 ± 0.066          & 0.755 ± 0.067          \\
                & Rad+Path+Gen      & 0.764 ± 0.062          & \textbf{0.788 ± 0.067} \\
Full Fusion     & Rad+Path+Gen+Clin & \textbf{0.775 ± 0.061} & 0.785 ± 0.077          \\ \bottomrule
\end{tabular}
\end{table}

\section{Results and Discussion}

Genomic- and pathology-only model performance metrics are practically similar (Table \ref{tab:table1}). However, the CNN-only (C-index=0.687 ± 0.067) and feature-only (C-index=0.653 ± 0.057) configurations of the radiology model underperform relative to the aforementioned unimodal models. Combining the radiology CNN features with the handcrafted features results in the strongest unimodal model. In contrast, clinical features are the least prognostic unimodal model.

\begin{figure}
    \centering
    \def\svgwidth{\columnwidth}
    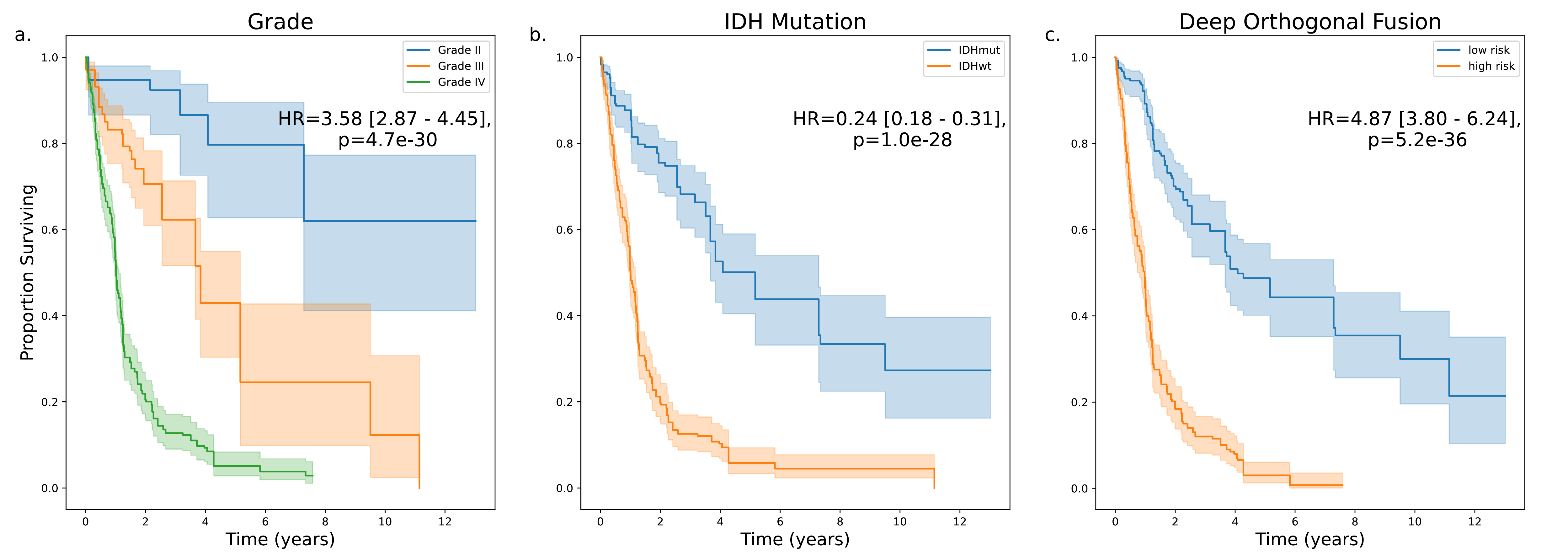
    \caption{Stratification by (a) grade, (b) IDH mutation status, and (c) DOF risk groups.} \label{fig3}
\end{figure}

 \begin{figure}
    \centering
    \def\svgwidth{\columnwidth}
    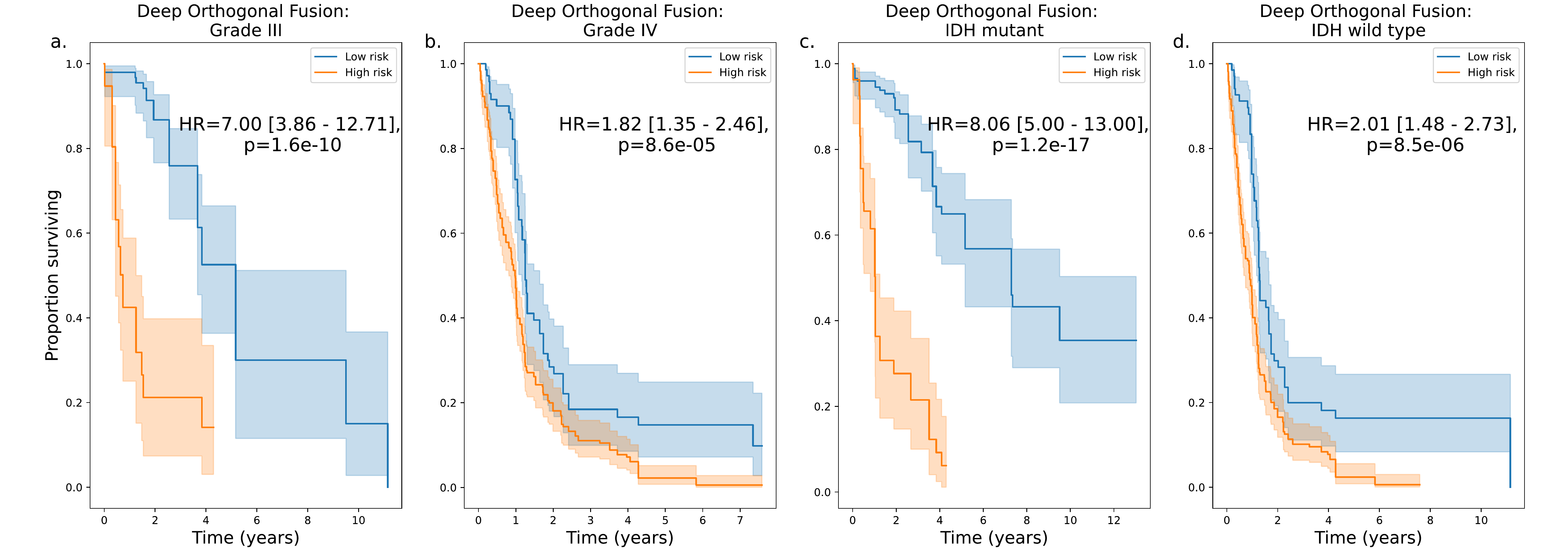
    \caption{DOF risk groups stratify patients by OS within (a,b) grade \& (c,d) IDH subsets.} \label{fig4}
\end{figure}

Deep fusion models integrating radiology outperform individual unimodal models, naive ensembles of unimodal models, as well as fusions of only clinical and/or biopsy-derived modalities.  The full fusion model (C-index=0.775 ± 0.061) achieves the best performance when trained with Cox loss \cite{ching_cox-nnet_2018} alone, second only to the Rad+Path+Gen model trained with MMO loss. Naive late fusion ensembles (i.e., averaging unimodal risk scores) exhibit inferior performance for Rad+Path+Gen with (C-index=0.735 ± 0.063) and without (C-index=0.739 ± 0.062) clinical features, confirming the benefits of deep fusion. 

The addition of MMO loss when training these deep fusion models consistently improves their performance at five different weightings (Table \ref{tab:tabs4}), with best performance for both at $\gamma=.5$. When all fusion models are trained at this weighting, 8 of 11 improve in performance. DOF combining radiology, pathology, and genomic data predicts glioma survival best overall with a median C-index of 0.788 ± 0.067, a significant increase over the best unimodal model (p=0.023). 

An ablation study was conducted to investigate the contributions of components of the fusion module (modality attention-gating and tensor fusion). We found that a configuration including both yields the best performance, but that strong results can also be achieved with a simplified fusion module (Table \ref{tab:tabs5}). 

In Fig. \ref{fig3}, KM plots show that the stratification of patients by OS in risk groups derived from this model perform comparably to established clinical markers. In Fig. \ref{fig4}, risk groups further stratify OS within grade and \textit{IDH} status groups. In sum, these results suggest that the DOF model provides useful prognostic value beyond existing clinical subsets and/or individual biomarkers. 

To further benchmark our approach, we implemented the fusion scheme of \cite{cheerla_deep_2019}, who combined pathology images, DNA, miRNA, and clinical data, which we further modified to also include radiology data. The network and learning approach is described in-depth in Table \ref{tab:tabs6}. In contrast to DOF, \cite{cheerla_deep_2019} instead seeks to maximize the correlation between modality embeddings prior to prediction. A model combining radiology, pathology, and genomic data achieved C-index=0.730 ± 0.05, while a model excluding the added radiology arm stratified patients by OS with C-index=0.715 ± 0.05.

\section{Conclusions}
We present DOF, a data efficient scheme for the novel fusion of radiology, histology, genomic, and clinical data for multimodal prognostic biomarker discovery. The integration of multi-dimensional data from biopsy-based modalities and radiology strongly boosts the ability to stratify glioma patients by OS. The addition of a novel MMO loss component, which forces unimodal embeddings to provide independent and complementary information to the fused prediction, further improves prognostic performance. Our DOF model incorporating radiology, histology, and genomic data significantly stratifies glioma patients by OS within outcome-associated subsets, offering additional granularity to routine clinical markers. DOF can be applied to any number of cancer domains, modality combinations, or new clinical endpoints including treatment response.

\printbibliography
\pagebreak
\section{Supplemental Information}
\beginsupplement

\begin{figure}[!hbt]
\centering
\includegraphics[width=.8\textwidth]{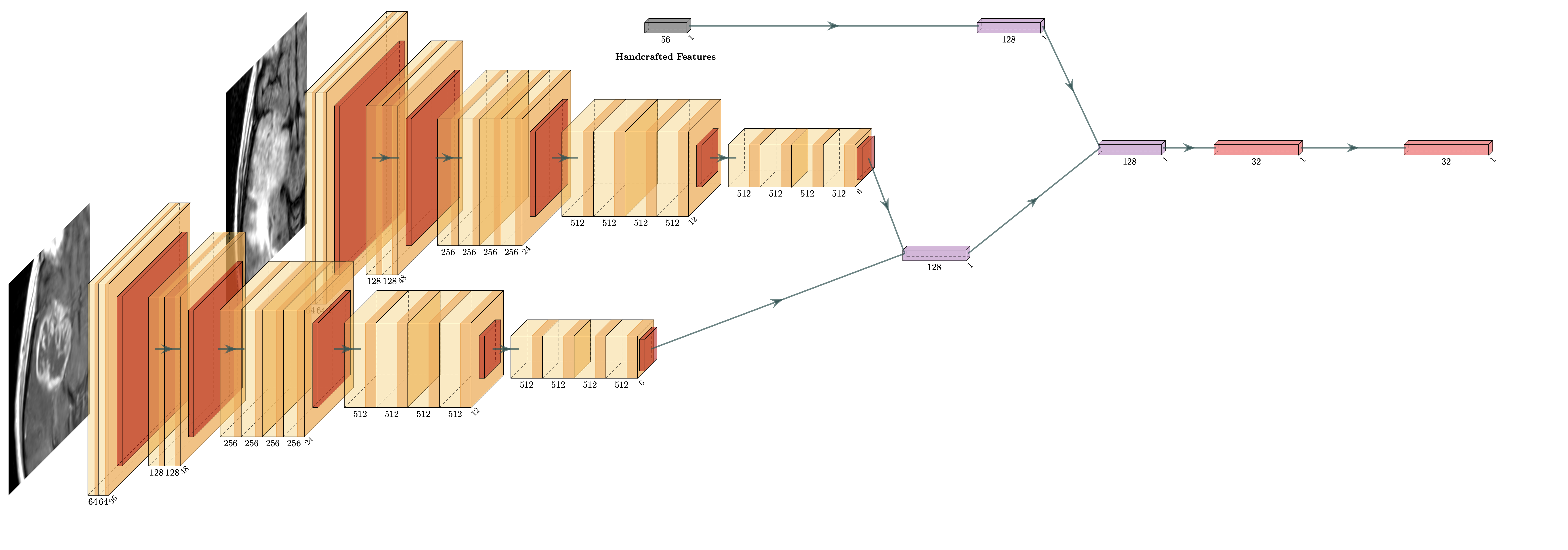}
\caption{Radiology FeatureNet combining images and features from Gd-T1w and T2w-FLAIR scans. }\label{figS1}
\end{figure}

  \begin{figure}[!hbt]
 \centering
\includegraphics[width=\textwidth]{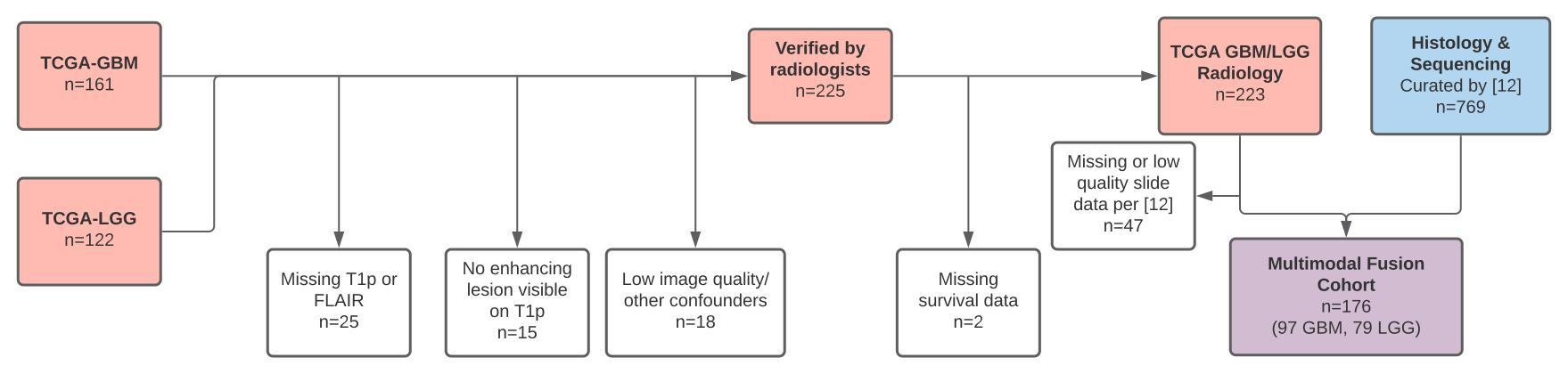}
\caption{Patient selection flowchart.} \label{figS2}
\end{figure}

\begin{table}[!hbt]
\caption{List of handcrafted radiology features. }
\label{tab:tabs1}
\resizebox{\textwidth}{!}{%
\begin{tabular}{@{}lll@{}}
\toprule
\textbf{Feature name/number} & \textbf{Feature Description}                    & \textbf{Summarization of annotated regions} \\ \midrule
No. regions (f1, f2)         & \# annotated lesions on Gd-T1w, edema on T2w-FLAIR      & N/A                                         \\
Volume (f3-f8)               & Volume of 3D ROI, measured in mm$^3$              & sum, largest, \& avg on Gd-T1w and T2w-FLAIR            \\
Longest axis (f9-f14)        & Longest distance between a contour’s vertices   & sum, largest, \& avg on Gd-T1w and T2w-FLAIR            \\
SA/V Ratio (f15-f20)         & Ratio of the surface area to volume.            & sum, largest, \& avg on Gd-T1w and T2w-FLAIR            \\
Sphericity (f21-f26)         & How closely a region’s shape resembles a sphere & sum, largest, \& avg on Gd-T1w and T2w-FLAIR            \\
Mean I (f27-f32)             & Mean intensity in contoured region              & sum, largest, \& avg on Gd-T1w and T2w-FLAIR            \\
10th percentile (f33-f38) & 10th \%  of intensities in contoured region & sum, largest, \& avg on Gd-T1w and T2w-FLAIR \\
90th percentile (f39-f44)    & 90th \% of intensities in contoured region      & sum, largest, \& avg on Gd-T1w and T2w-FLAIR            \\
Skewness (f45-f50)           & Skewness of intensities in contoured region     & sum, largest, \& avg on Gd-T1w and T2w-FLAIR            \\
Variance (f51-f56)           & Variance of intensities in contoured region     & sum, largest, \& avg on Gd-T1w and T2w-FLAIR            \\ \bottomrule
\end{tabular}%
}
\end{table}

\begin{table}[]
\tiny
\caption{List of molecular features}
\label{tab:tabs2}
\resizebox{\textwidth}{!}{%
\begin{tabularx}{\textwidth}{lXl}
\toprule
\textbf{Category} & \textbf{Variable}        & \textbf{Value Type}                                \\
\midrule
Gene-level CNV (f1-f41) & 
  EGFR, MDM4, MYC, BRAF, EZH2, MET, SMO, KIAA1549, CREB3L2, NTRK1, PRCC, BLM, NTRK3, CRTC3, CDKN2A, CDKN2B, FGFR2, TSHR, TCL1A, TRIP11, GOLGA5, GPHN, DICER1, TCL6, EBF1, ITK, RPL22, CDKN2C, LCP1, RB1, IRF4, FGFR1OP, MLLT4, MYB, ROS1, TNFAIP3, GOPC, CARD11, JAK2, STK11, PTEN  & Categorical\\
Arm-level CNV (f42-f78) &
  1q, 2p, 2q, 3p, 3q, 4p, 4q, 5p, 5q, 6p, 6q, 7p, 7q, 8p, 8q, 9p, 9q, 10p, 10q, 11p, 11q, 12p, 12q, 13q, 14q, 15q, 16p, 16q, 17p, 17q, 18p, 18q, 19p, 20p, 20q, 21q, 22q & Continuous \\
Biomarkers (f79, f80)     & IDH Mutation, 1p/19q Codeletion & Binary \\
\bottomrule
\end{tabularx}%
}
\end{table}

\begin{table}[!hbt]
\caption{List of clinical features. }
\label{tab:tabs3}
\resizebox{\textwidth}{!}{%
\begin{tabular}{@{}ll@{}}
\toprule
\textbf{Variable}                                      & \textbf{Value Type} \\ \midrule
Age (f1)                                               & Continuous          \\
Karnofsky Performance Score (f2)                       & Continuous          \\
Grade (f3)                                             & Categorical         \\
Sex: Male vs. Female (f4)                              & Binary              \\
Treatment: any (f5), radiation (f6), chemotherapy (f7) & Binary              \\
Histological diagnosis: LGG (f8), Astrocytoma (f9), Glioblastoma (f10),  Oligoastrocytoma (f11), Oligodendroglioma (f12) & Binary \\
Race/ethnicity: White vs. Non-white (f13), Hispanic vs. Non-hispanic (f14)                                               & Binary \\ \bottomrule
\end{tabular}%
}
\end{table}

\begin{table}[]
\small
\centering
\caption{Median C-index for fusion models at various MMO loss weightings.}
\label{tab:tabs4}
\begin{tabular}{@{}ccc@{}}
\toprule 
\textbf{$\gamma$} & \textbf{  Rad + Path + Gen  }    & \textbf{  Rad + Path + Gen + Clin  } \\ \midrule
0                     & 0.764 ± 0.062 & 0.775 ± 0.061   \\
.1                    & 0.768 ± 0.064 & 0.745 ± 0.068   \\
.25                   & 0.777 ± 0.066 & 0.782 ± 0.066   \\
.5                    & 0.788 ± 0.067 & 0.785 ± 0.077   \\
1                     & 0.779 ± 0.070 & 0.776 ± 0.075   \\
2.5                   & 0.781 ± 0.073 & 0.760 ± 0.072   \\ \bottomrule
\end{tabular}
\end{table}

\begin{table}[]
\centering
\small
\caption{Ablation study investigating impact of components of fusion module for best-performing modality combination (Rad + Path + Gen)}
\label{tab:tabs5}
\resizebox{7.25cm}{!}{%
\begin{tabular}{@{}lll@{}}
\toprule
\textbf{  Attention Gating  } & \textbf{  Combination Strategy  } & \textbf{  Median C-index  } \\
\midrule
Yes              & Tensor Fusion        & 0.79 ± 0.07    \\
No               & Tensor Fusion        & 0.77 ± 0.08    \\
Yes              & Concatenation        & 0.78 ± 0.07    \\
No               & Concatenation        & 0.76 ± 0.07   \\

\end{tabular}%
}
\end{table}

\begin{table}[]

\begin{minipage}{\linewidth}
\tiny
\caption{Correlation-based deep fusion framework \cite{cheerla_deep_2019}, adapted to include radiology. }
\label{tab:tabs6}
\resizebox{\textwidth}{!}{%
\begin{tabularx}{\textwidth}{lX}
\toprule
\textbf{Module} & \textbf{Description}                                                                                                                                                                                                                                                                         \\ \midrule
DNA                                                   & Fully connected (FC) layer. Unmodified from \cite{cheerla_deep_2019}.                                                                                                                                                                                                                                                   \\
Pathology                                            & Squeezenet applied to histology ROIs, followed by a FC layer. While \cite{cheerla_deep_2019} trained from scratch, we found better results when using pretrained ImageNet weights and freezing the convolutional layers                                    \\
Radiology                                             & Not included in \cite{cheerla_deep_2019}. We applied pre-trained, frozen squeezenet to Gd-T1w and T2w-FLAIR ROIs. FC layers were applied to CNN-extracted features and radiomics features, yielding 128 features from each. These were concatenated and processed by another FC layer \\
Fusion                                          & The output of each unimodal arm is a feature vector of length 256, which were averaged together. The averaged feature representation is then processed by a 10-layer highway network. Unmodified from \cite{cheerla_deep_2019}                                                                                                \\
Loss                                                     & Combination of Cox proportional hazard-based loss for prognosis prediction and similarity loss that enforces correlated representations between modalities. Unmodified from \cite{cheerla_deep_2019}.                                                                                                   \\ \bottomrule

\end{tabularx}%
}

\par
\tiny
*All changes from \cite{cheerla_deep_2019} made by us are noted specifically above. Any further differences from the description of \cite{cheerla_deep_2019} are due to discrepancies between the paper and its codebase.\end{minipage}

\end{table}
\end{document}